\documentclass{article}

\usepackage{PRIMEarxiv}

\usepackage[utf8]{inputenc} 
\usepackage[T1]{fontenc}    
\usepackage{hyperref}       
\usepackage{url}            
\usepackage{booktabs}       
\usepackage{amsfonts}       
\usepackage{nicefrac}       
\usepackage{microtype}      
\usepackage{lipsum}
\usepackage{fancyhdr}       
\usepackage{graphicx}       
\graphicspath{{media/}}     

\title{FoodGPT: A Large Language Model in Food Testing Domain with Incremental Pre-training and Knowledge Graph Prompt}

\author{
  Zhixiao Qi, Yijiong Yu, Meiqi Tu \\
  Department of Automation \\
  Tsinghua University \\
  Beijing\\
  \texttt{\{qzx21, yyj22, tmq21\}@mails.tsinghua.edu.cn} \\
   \And
  Junyi Tan \\
  School of Information and Communication Engineering \\
  Beijing University of Posts and Telecommunications \\
  Beijing\\
  \texttt{emilytamjam@outlook.com} \\
   \And
  Yongfeng Huang \\
  Department of Electronic Engineering \\
  Tsinghua University \\
  Beijing\\
  \texttt{yfhuang@tsinghua.edu.cn} \\
}

\begin{document}
\maketitle

\begin{abstract}
Currently, the construction of large language models in specific domains is done by fine-tuning on a base model. Some models also incorporate knowledge bases without the need for pre-training. This is because the base model already contains domain-specific knowledge during the pre-training process. We build a large language model for food testing. Unlike the above approach, a significant amount of data in this domain exists in Scanning format for domain standard documents. In addition, there is a large amount of untrained structured knowledge. Therefore, we introduce an incremental pre-training step to inject this knowledge into a large language model. In this paper, we propose a method for handling structured knowledge and scanned documents in incremental pre-training. To overcome the problem of machine hallucination, we constructe a knowledge graph to serve as an external knowledge base for supporting retrieval in the large language model. It is worth mentioning that this paper is a technical report of our pre-release version, and we will report our specific experimental data in future versions.
\end{abstract}


\section{Introduction}
Large language models (LLM) \cite{brown2020language} have gained significant research importance in the field of natural language processing. Models such as ChatGPT, LLaMA \cite{touvron2023llama}, GPT-4, ChatGLM \cite{du2021glm}, and PaLM \cite{chowdhery2022palm} have demonstrated outstanding performance in downstream tasks. The powerful ability of LLM in understanding human instructions has led to continuous research on LLMs in various vertical domains.

ChatLaw \cite{cui2023chatlaw} is based on Ziya-LLaMA-13B and utilizes legal data for instruction fine-tuning, incorporating vector database retrieval to create a legal LLM. DoctorGLM \cite{xiong2023doctorglm} is built upon ChatGLM-6B and fine-tuned using Chinese medical dialogue datasets to create a Chinese medical consultation model. BenTsao is based on LLaMA-7B and constructs a Chinese medical LLM by leveraging a medical knowledge graph and the GPT-3.5 API to build a Chinese medical instruction dataset. Cornucopia, on the other hand, is based on LLaMA-7B and constructs an instruction dataset using Chinese financial public data and crawled financial data, focusing on question-answering in the financial domain.

Previous research assume that the base models have already injected the corresponding domain knowledge, hence no incremental pre-training is performed on the base models. However, in certain domains, such as food testing, a large amount of knowledge exists in non-textual images, scanned documents and private structured data. To build LLMs in these domains, we believe that the knowledge in the base models is insufficient, making incremental pre-training essential. Therefore, we begin constructing a Chinese food testing LLM from the incremental pre-training stage, using the food testing domain as a representative example. 

In summary, this paper contributes in the following ways:
\begin{itemize}
\item We develope a LLM called FoodGPT for food testing, which, to our knowledge, is the first LLM in the field of food testing.
\item We start building vertical domain LLM from incremental pre-training, which is currently a rare effort to our knowledge.
\item For scanned format domain standard documents and structured data, we propose a new data collection method that successfully injects the above knowledge into the FoodGPT during the incremental pre-training stage.
\item The food testing domain LLM requires precise numerical outputs. Hence, we construct a knowledge graph as a retrieval library for the FoodGPT to reduce machine hallucination phenomena.
\end{itemize}

\section{Incremental Pre-training}
In the field of food testing, a significant amount of data is present in images or scanned documents, while another portion is stored in private structured databases. These data have not been utilized in the pre-training of LLMs. If we directly fine-tune the base model, it may struggle to perform well due to a lack of domain knowledge. Therefore, in the training of FoodGPT, we incorporate an incremental pre-training step to inject domain knowledge into FoodGPT. Below, we will describe the approach to handling different types of pre-training data.

\begin{figure*}[htp]
    \centering
    \includegraphics[width=1.0\textwidth]{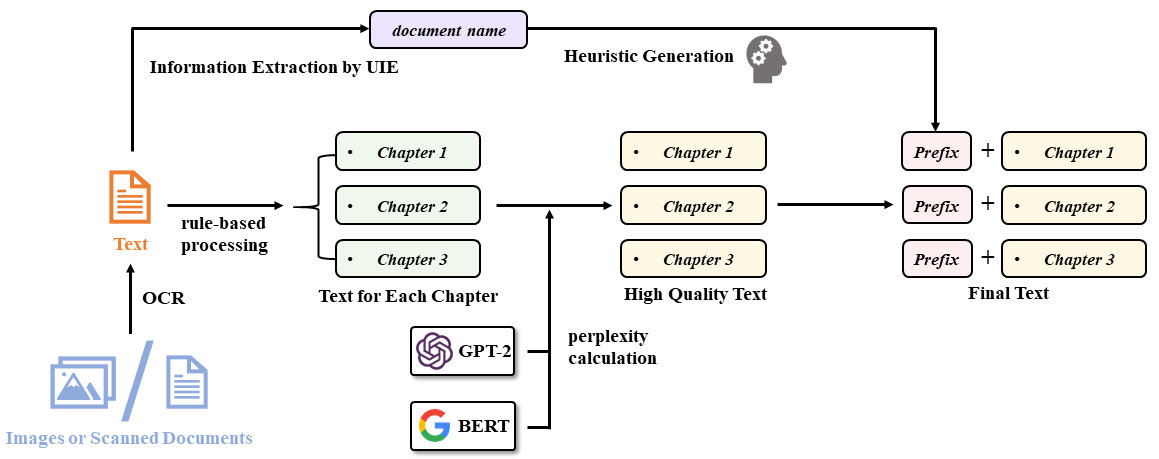}
    \caption{Processing pipeline of images and scanned documents.}
    \label{fig:one}
\end{figure*}

\subsection{Images and Scanned Documents}
The storage of images and scanned documents contains a majority of domain standard document information. We categorize these two types of data together because we use optical character recognition (OCR) technology to process both. Due to the inability to directly extract text from these images and scanned documents, we apply OCR to over ten thousand of them. The specific processing pipeline is illustrated in Figure \ref{fig:one}. Each document represents a domain standard document and exceeds the maximum sequence length used for training the model. Therefore, we split the documents into chapters based on their sections. It is discovered that descriptions of the same testing item varied among different documents, as different food items are targeted by the same testing item in different documents. To prevent conflicts in these descriptions, we add a prefix before each chapter of data to indicate the corresponding document. We fine-tune a UIE \cite{lu2022unified} model to extract document names from text, and construct the prefixes for the extracted document names using a heuristic generation method, which are then concatenated with the text. Additionally, each chapter of text may include data such as tables and formulas. These types of data, when processed by OCR, significantly affect the fluidity of the text. Therefore, we used BERT \cite{devlin2018bert} and GPT-2 \cite{radford2019language} to calculate the perplexity of each sentence in each chapter of text, and exclude sentences with high perplexity.

\subsection{Structured Knowledge}
\begin{figure*}[htp]
    \centering
    \includegraphics[width=1.0\textwidth]{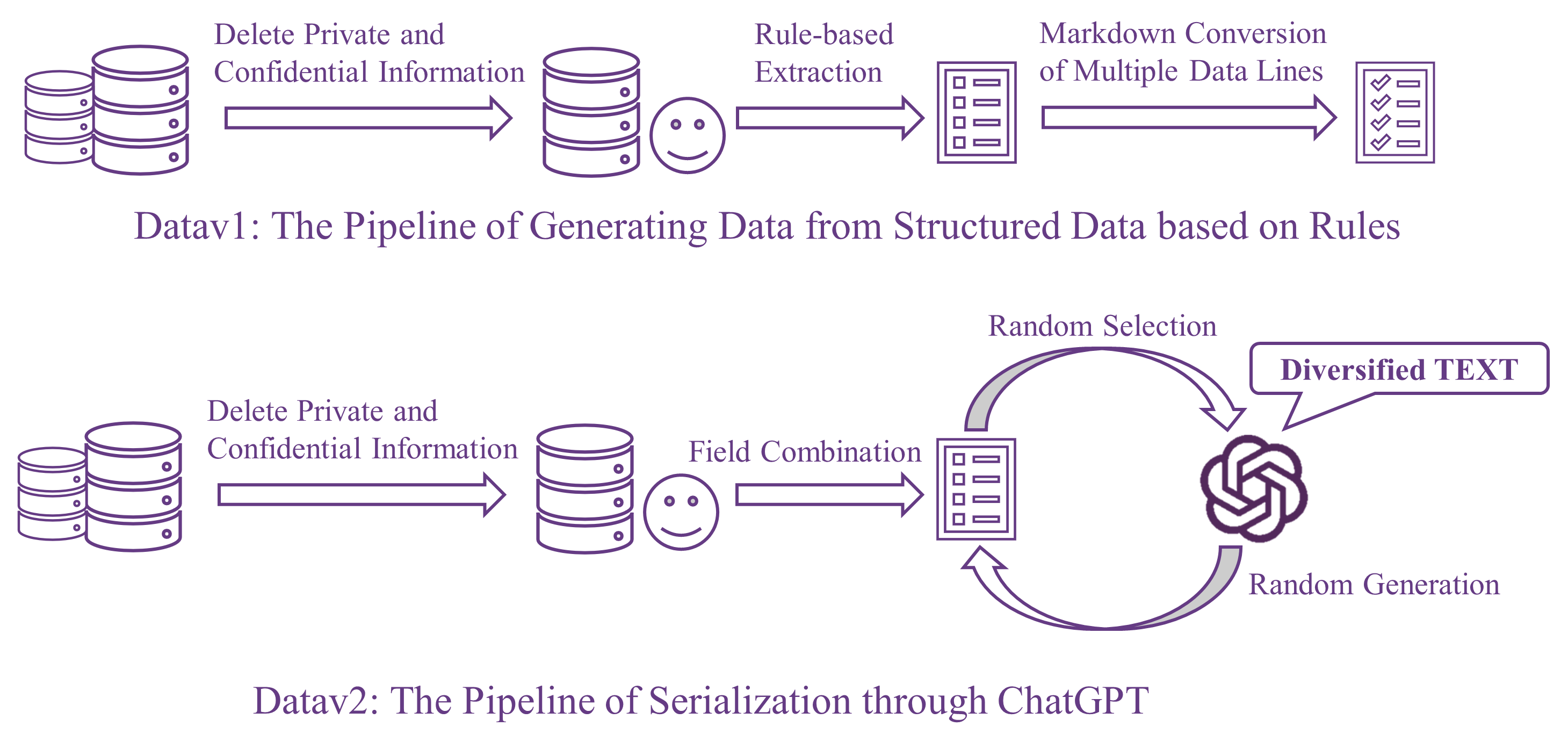}
    \caption{There are two ways we handle structured data.}
    \label{fig:two}
\end{figure*}

A large amount of knowledge in the field of food testing is also present in private structured databases. These data consist of manually entered tables obtained from extensive food testing and are stored in private databases of testing institutions or intermediaries. They contain a wealth of learnable knowledge. In addition to integrating this knowledge into an external knowledge base, we believe that it should also be incorporated into the incremental pre-training of LLM. The way we handle structured data in incremental pre-training is illustrated in Figure \ref{fig:two}.

We create two versions of structured data, Datav1 and Datav2, for incremental pre-training. In Datav1, we first remove the confidential and privacy information fields from the table. Then, we construct each data by \textit{dict}. It's worth noting that in this domain, one food item can correspond to multiple testing items. We use the "testing item" as the \textit{key} and represent the corresponding multiple specific testing items using a table in markdown format as the \textit{value}.

We consider that this structured data could potentially harm the model, so it is necessary to serialize the structured knowledge. As far as we know, currently structured data is manually converted into natural language using templates. However, the number of manual templates is quite limited, and there may be ambiguous repetitions and a decrease in model performance when serializing a large amount of structured knowledge. Therefore, our Datav2 employs a novel method to serialize structured data. The construction process of Datav2 is as follows:

\begin{itemize}
\item Similar to Datav1, we remove confidential and private information fields from the table.
\item Because some fields in the original data do not have individual meanings, we merge certain fields to ensure each field has a separate meaning.
\item We input each data point into ChatGPT to generate text randomly according to specific rules.
The rules for random text generation are as follows: 1) Randomly select fields to input for each data point. 2) Randomly choose the temperature parameter within the range of [0.5, 1.0] for text generation. 3) Ensure that each field in a data point is selected at least once and no more than twice during text generation, in order to include all field information and avoid excessive repetition.
\end{itemize}

\subsection{Other Types of Data}
In addition to the two types of data mentioned above, we incorporate other data to construct the incremental pre-training dataset for FoodGPT. The specific sources are as follows:
\begin{itemize}
\item Food detection dictionary. Each data entry provides an explanation of specialized terms.
\item Chinese tutorials and research papers in the field of food testing. We split them into individual data entries based on paragraphs.
\item Food sentiment data. Each data entry consists of news articles related to public opinions on food, covering the past five years.
\item Food safety-related laws. Each data entry represents a provision within the laws.
\item Food safety-related exam questions. Each data entry corresponds to a question, and we make efforts to include detailed explanations for each question.
\end{itemize}

We choose Chinese-LLaMA2-13B as the base model and used the LoRA \cite{hu2021lora} method for incremental pre-training. We will elaborate on the specific experimental results in future versions of the technical report.

\begin{figure*}[htp]
    \centering
    \includegraphics[width=0.7\textwidth]{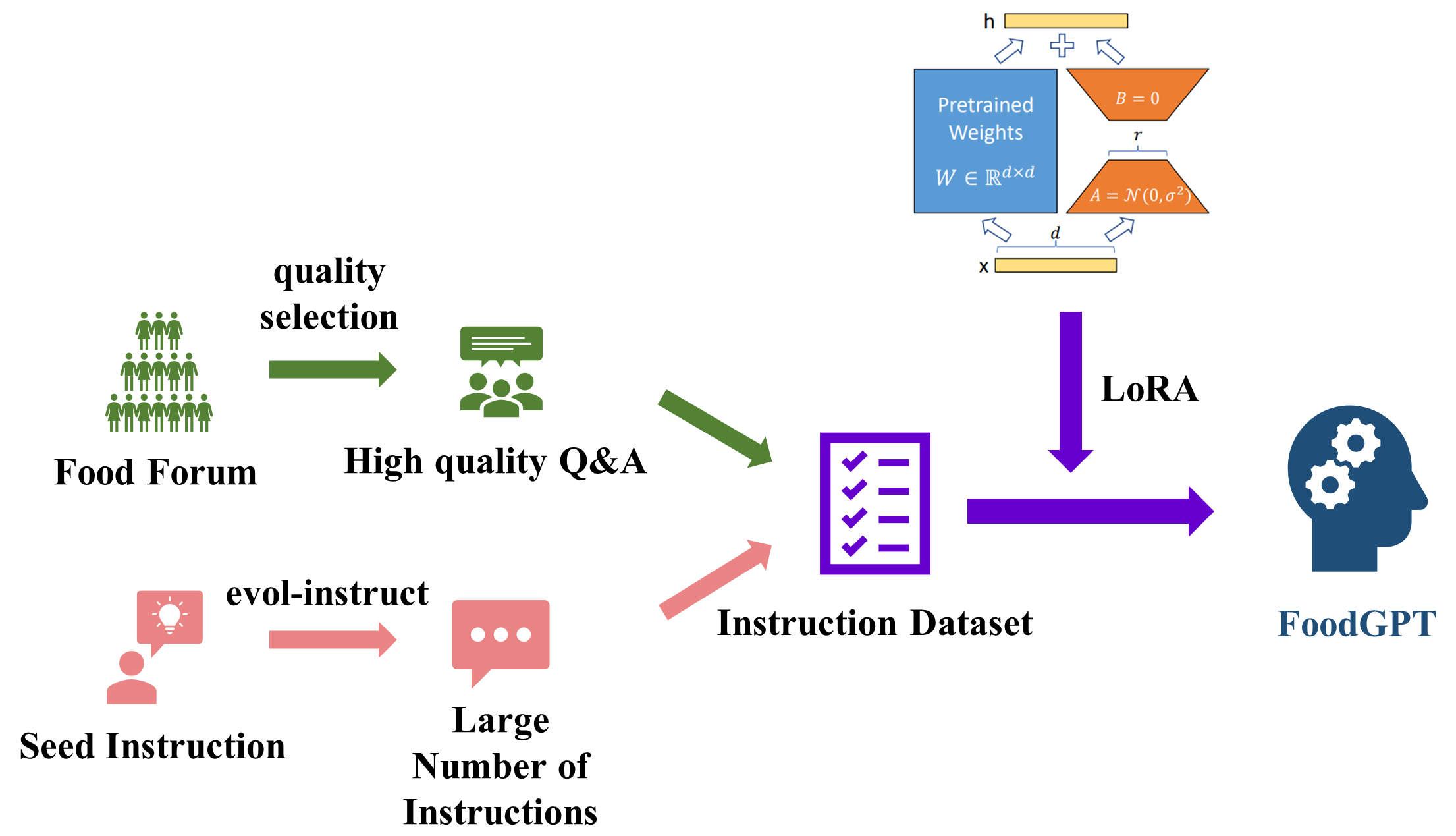}
    \caption{The entire pipeline of instruction fine-tuning.}
    \label{fig:three}
\end{figure*}

\section{Instruction Fine-tuning}
During the instruction fine-tuning phase, we construct a dataset for instruction fine-tuning through two channels. The LoRA approach was employed to fine-tune FoodGPT.
\subsection{Dataset}
We utilize two methods to construct the instruction fine-tuning dataset for FoodGPT. Firstly, we select relevant topics from food forums and scraped a large number of question-answer pairs. To ensure high-quality answers, we prioritize users with higher posting frequency, aiming to include answers from users who frequently post quality content. Secondly, we collaborate with industry experts in the food testing field to design 100 high-quality seed instructions. These seed instructions are further expanded and diversified using the evol-instruct \cite{xu2023wizardlm} method, resulting in a more comprehensive and extensive dataset of instructions.
\subsection{Training Process}
We utilize the LoRA method to fine-tune the instructions for Chinese-LLaMA2-13B. The entire pipeline of instruction fine-tuning is illustrated in Figure \ref{fig:three}.

\section{External Knowledge Graph Retrieval}
\begin{figure*}[htp]
    \centering
    \includegraphics[width=0.5\textwidth]{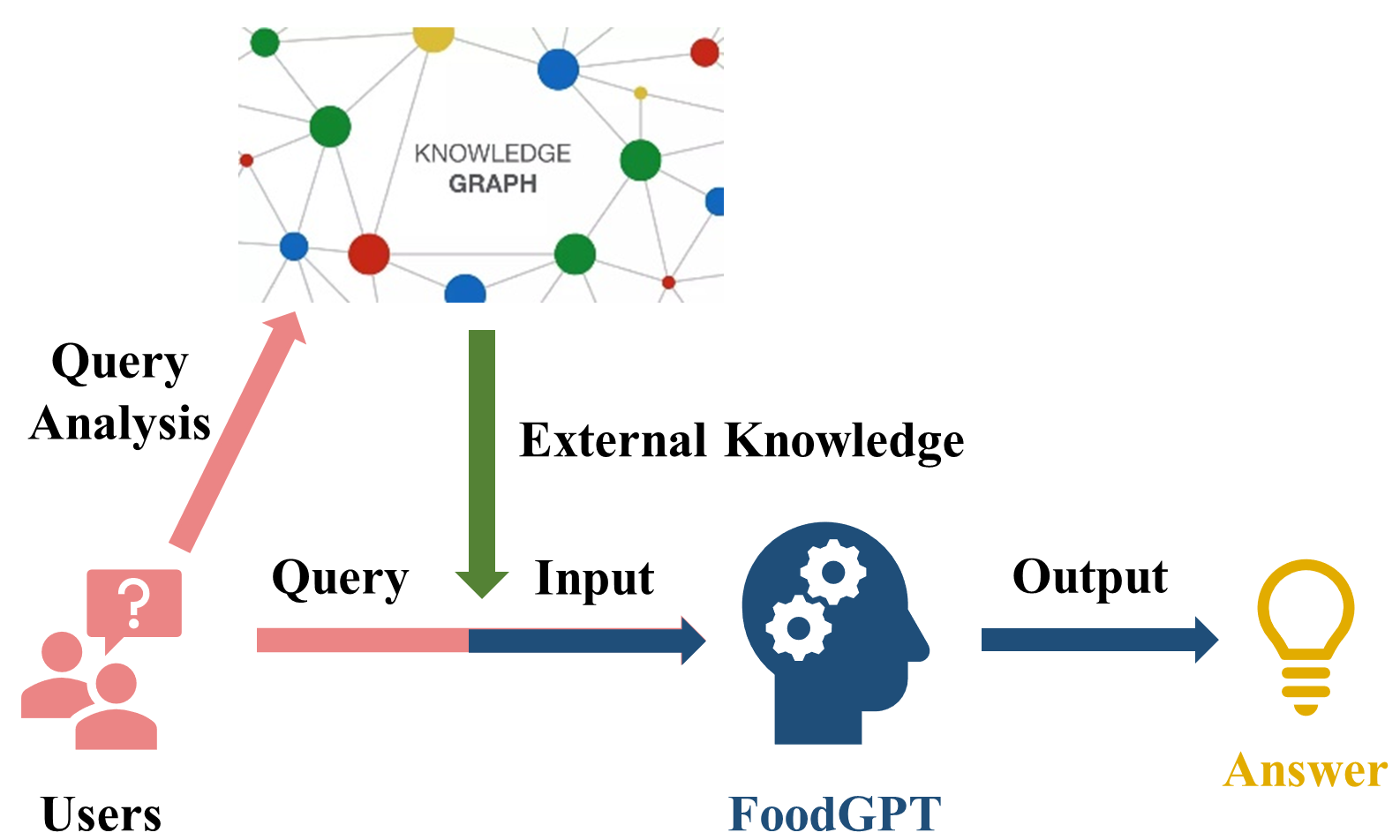}
    \caption{Retrieval Method for External Knowledge Graph.}
    \label{fig:four}
\end{figure*}

The requirements for food testing are very strict in terms of output. For instance, inaccurate data output could lead to severe consequences. In order to ensure the quality of the data generated by FoodGPT and reduce machine hallucinations, we construct a knowledge graph by extracting a large amount of knowledge from structured data and text. This knowledge graph serves as an external knowledge base to support the retrieval and output of FoodGPT. The process of incorporating the knowledge graph is depicted in Figure \ref{fig:four}. When the user inputs a query to FoodGPT, we utilize a retrieval model to parse the query and retrieve relevant knowledge from the knowledge graph. The retrieved knowledge is then concatenated with the query and inputted into FoodGPT. FoodGPT, equipped with parameterized knowledge and external knowledge, comprehends the user's intent and generates responses.

\section{Conclusion}
In this paper, we present FoodGPT, a large language model for the field of food testing. FoodGPT is built upon the Chinese-LLaMA2-13B base model, with incremental pre-training, instruction fine-tuning, and external knowledge graph integration. Since a significant amount of knowledge exists in images, scanned documents, and private structured knowledge bases, which the base model lacks, we deem it necessary to perform incremental pre-training. We propose new approaches to handle these data and incorporate them into our incremental pre-training database along with other data. In the instruction fine-tuning stage, we crawl question-answer pairs from forums and employ evol-instruct to construct the fine-tuning dataset, guided by seed instructions provided by domain experts. Given the stringent requirements for output metrics in the field of food testing, we additionally construct a knowledge graph to serve as an external database to assist FoodGPT in generating outputs.

It is worth mentioning that this paper is a technical report of the pre-release version of FoodGPT, and we will elaborate on experimental details and analysis in future versions.

\newpage
\bibliographystyle{unsrt}  
\bibliography{references}

\begin{thebibliography}{10}

\bibitem{brown2020language}
Tom Brown, Benjamin Mann, Nick Ryder, Melanie Subbiah, Jared~D Kaplan, Prafulla
  Dhariwal, Arvind Neelakantan, Pranav Shyam, Girish Sastry, Amanda Askell,
  et~al.
\newblock Language models are few-shot learners.
\newblock {\em Advances in neural information processing systems},
  33:1877--1901, 2020.

\bibitem{touvron2023llama}
Hugo Touvron, Thibaut Lavril, Gautier Izacard, Xavier Martinet, Marie-Anne
  Lachaux, Timoth{\'e}e Lacroix, Baptiste Rozi{\`e}re, Naman Goyal, Eric
  Hambro, Faisal Azhar, et~al.
\newblock Llama: Open and efficient foundation language models.
\newblock {\em arXiv preprint arXiv:2302.13971}, 2023.

\bibitem{du2021glm}
Zhengxiao Du, Yujie Qian, Xiao Liu, Ming Ding, Jiezhong Qiu, Zhilin Yang, and
  Jie Tang.
\newblock Glm: General language model pretraining with autoregressive blank
  infilling.
\newblock {\em arXiv preprint arXiv:2103.10360}, 2021.

\bibitem{chowdhery2022palm}
Aakanksha Chowdhery, Sharan Narang, Jacob Devlin, Maarten Bosma, Gaurav Mishra,
  Adam Roberts, Paul Barham, Hyung~Won Chung, Charles Sutton, Sebastian
  Gehrmann, et~al.
\newblock Palm: Scaling language modeling with pathways.
\newblock {\em arXiv preprint arXiv:2204.02311}, 2022.

\bibitem{cui2023chatlaw}
Jiaxi Cui, Zongjian Li, Yang Yan, Bohua Chen, and Li~Yuan.
\newblock Chatlaw: Open-source legal large language model with integrated
  external knowledge bases.
\newblock {\em arXiv preprint arXiv:2306.16092}, 2023.

\bibitem{xiong2023doctorglm}
Honglin Xiong, Sheng Wang, Yitao Zhu, Zihao Zhao, Yuxiao Liu, Qian Wang, and
  Dinggang Shen.
\newblock Doctorglm: Fine-tuning your chinese doctor is not a herculean task.
\newblock {\em arXiv preprint arXiv:2304.01097}, 2023.

\bibitem{lu2022unified}
Yaojie Lu, Qing Liu, Dai Dai, Xinyan Xiao, Hongyu Lin, Xianpei Han, Le~Sun, and
  Hua Wu.
\newblock Unified structure generation for universal information extraction.
\newblock {\em arXiv preprint arXiv:2203.12277}, 2022.

\bibitem{devlin2018bert}
Jacob Devlin, Ming-Wei Chang, Kenton Lee, and Kristina Toutanova.
\newblock Bert: Pre-training of deep bidirectional transformers for language
  understanding.
\newblock {\em arXiv preprint arXiv:1810.04805}, 2018.

\bibitem{radford2019language}
Alec Radford, Jeffrey Wu, Rewon Child, David Luan, Dario Amodei, Ilya
  Sutskever, et~al.
\newblock Language models are unsupervised multitask learners.
\newblock {\em OpenAI blog}, 1(8):9, 2019.

\bibitem{hu2021lora}
Edward~J Hu, Yelong Shen, Phillip Wallis, Zeyuan Allen-Zhu, Yuanzhi Li, Shean
  Wang, Lu~Wang, and Weizhu Chen.
\newblock Lora: Low-rank adaptation of large language models.
\newblock {\em arXiv preprint arXiv:2106.09685}, 2021.

\bibitem{xu2023wizardlm}
Can Xu, Qingfeng Sun, Kai Zheng, Xiubo Geng, Pu~Zhao, Jiazhan Feng, Chongyang
  Tao, and Daxin Jiang.
\newblock Wizardlm: Empowering large language models to follow complex
  instructions.
\newblock {\em arXiv preprint arXiv:2304.12244}, 2023.

\end{thebibliography}

\end{document}